\documentclass[letterpaper, 10 pt, conference]{ieeeconf}  

\IEEEoverridecommandlockouts                              

\usepackage{graphicx}
\usepackage{multirow}
\usepackage{amssymb}
\usepackage{amsmath}
\usepackage{amstext}
\usepackage{comment}
\usepackage[dvipsnames]{xcolor}
\usepackage{pifont}
\usepackage{array}
\newcommand{\PreserveBackslash}[1]{\let\temp=\\#1\let\\=\temp}
\newcolumntype{C}[1]{>{\PreserveBackslash\centering}p{#1}}
\newcolumntype{R}[1]{>{\PreserveBackslash\raggedleft}p{#1}}
\newcolumntype{L}[1]{>{\PreserveBackslash\raggedright}p{#1}}
\usepackage[colorlinks=true,linkcolor=blue,urlcolor=blue]{hyperref}

\usepackage{textcomp}
\newcommand{\ci}[1]{\tiny{\textcolor{gray}{~($\pm #1$)}}}
\definecolor{commentcolor}{HTML}{0071bc}

\definecolor{citecolor}{HTML}{3953A4} 
\definecolor{ogreen}{HTML}{2E7D32}
\definecolor{bred}{HTML}{BF360C}
\definecolor{newbrown}{HTML}{795548}

\usepackage{hyperref}
\makeatletter
\let\NAT@parse\undefined
\makeatother
\usepackage[numbers,sort,compress]{natbib}
\usepackage[all=normal,bibbreaks=tight,paragraphs=tight,floats=tight]{savetrees}
\hypersetup{
colorlinks=true, linkcolor=bred, urlcolor=magenta, citecolor=citecolor,
}
\usepackage[linewidth=1pt, linecolor=blue]{mdframed}
 \usepackage{url}

\title{\LARGE \bf
Human-in-the-loop Embodied Intelligence with \\ Interactive Simulation Environment for Surgical Robot Learning
}

\author{
Yonghao Long, Wang Wei, Tao Huang, Yuehao Wang and Qi Dou \\
The Chinese University of Hong Kong
\thanks{This project was supported in part by Hong Kong Research Grants Council TRS Project No.T42-409/18-R, in part by Hong Kong Innovation and Technology Commission under Project No. PRP/026/22FX, in part by Multi-Scale Medical Robotics Center InnoHK under grant 8312051, and in part by a Research Fund from Cornerstone Robotics Ltd.}
\thanks{Y. Long, W. Wei, T. Huang, Y. Wang, and Q. Dou are with the Department of Computer Science and Engineering, The Chinese University of Hong Kong. Q. Dou is also with the T Stone Robotics Institute, CUHK.}
\thanks{\textit{Corresponding author: Qi Dou (qidou@cuhk.edu.hk).}}%
}
\begin{document}
\maketitle
\thispagestyle{empty}
\pagestyle{empty}
\begin{abstract}
Surgical robot automation has attracted increasing research interest over the past decade, expecting its potential to benefit surgeons, nurses and patients. Recently, the learning paradigm of embodied intelligence has demonstrated promising ability to learn good control policies for various complex tasks, where embodied AI simulators play an essential role to facilitate relevant research. 
However, existing open-sourced simulators for surgical robot are still not sufficiently supporting human interactions through physical input devices, which further limits effective investigations on how the human demonstrations would affect policy learning.
In this work, we study human-in-the-loop embodied intelligence with a new interactive simulation platform for surgical robot learning. Specifically, we establish our platform based on our previously released SurRoL simulator with several new features co-developed to allow high-quality human interaction via an input device. 
We showcase the improvement of our simulation environment with the designed new features, and validate effectiveness of incorporating human factors in embodied intelligence through the use of human demonstrations and reinforcement learning as a representative example. Promising results are obtained in terms of learning efficiency. Lastly, five new surgical robot training tasks are developed and released, with which we hope to pave the way for future research on surgical embodied intelligence. 
Our learning platform is publicly released and will be continuously updated in the website: \url{https://med-air.github.io/SurRoL}.
\end{abstract}
\section{Introduction}
Surgical robotics has developed rapidly in the past decade and transformed minimally invasive surgery in practice~\cite{taylor2022surgical}. Recently, surgical task automation~\cite{d2021accelerating} has received increasing attention from researchers as it is promising to reduce burden of surgeons and improve operational efficiency~\cite{yang2018grand}. However, it still remains a distant dream with challenges from complex surgical scenes and skillful surgical actions.
Conventional solutions typically rely on heuristic planning methods while struggle to yield scalable control policies. 
To date, successful stories on surgical task automation are still in its infancy~\cite{maier2022surgical}.


Embodied intelligence~\cite{duan2022survey} hypothesizes to directly learn various skills based on interactions between robots and the environment, which has demonstrated remarkable capability on robot task automation~\cite{savva2019habitat}. 
In this context, simulators~\cite{ramrakhya2022habitat,li2021igibson,Xiang_2020_SAPIEN} serve as the essential infrastructure to provide a digital twin~\cite{bonne2022digital,shu2022twin} of the physical world, in order to facilitate collection of interactive data, training and testing of the agent. Reinforcement learning (RL)~\cite{ibarz2021train} is typically used together with embodied AI, which can model the task execution as Markov Decision Process and optimize the policy learning through robot-environment interactions in the simulator. Despite visible efforts that have been made on embodied intelligence in general~\cite{ramrakhya2022habitat,ai2thor,li2021igibson}, surgical embodied intelligence, which should be supported by tailored and domain-specific simulation environments still remains to be further explored.

\begin{figure}[t]
    \centering
    \vspace{0.2cm}
    \includegraphics[width=0.95\hsize]{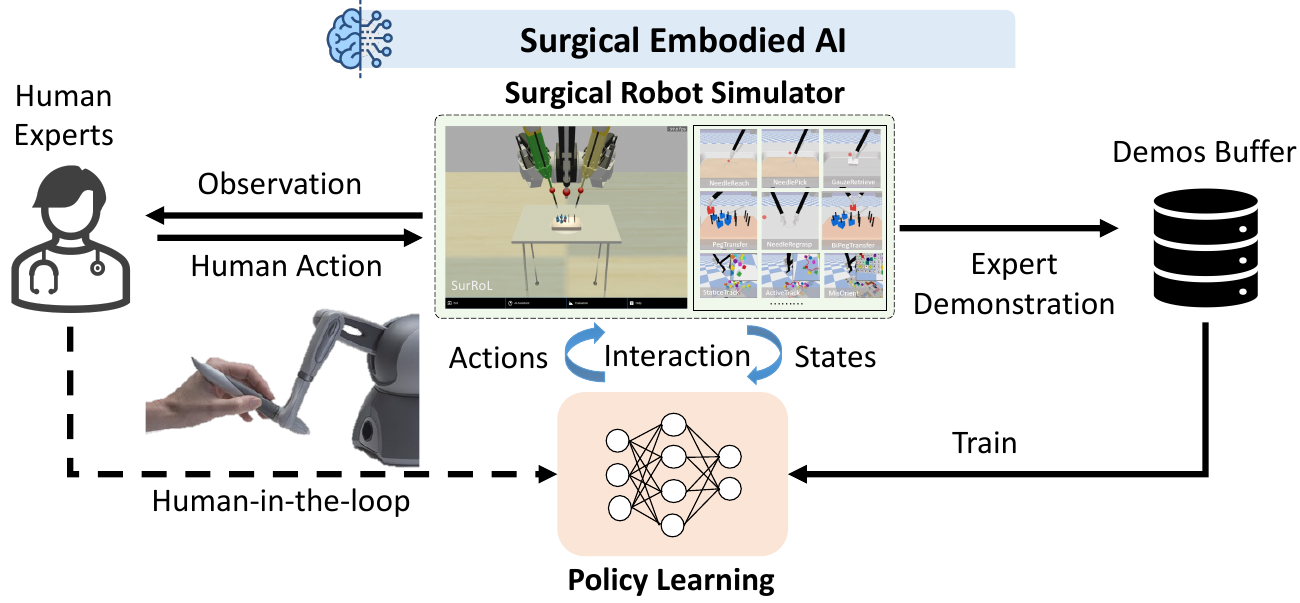}
    \vspace{-3mm}
    \caption{Illustration for the concept of surgical embodied intelligence with human-in-the-loop demonstrations for robot learning.}
    \label{TrainingFramework}
    \vspace{-7mm}
\end{figure}

Specifically, early works~\cite{kenney2009face,whittaker2016validation} focusing on simulation of surgical robot and its relevant tasks (such as peg transfer) were concentrated on realistic virtual environment creation, rather than serving for AI purposes as is desired nowadays.
Recent simulators increasingly aim at bridging embodied intelligence, in particular reinforcement learning, to surgical robots for developing advanced methods for task automation. 
Richter \emph{et al.} proposed dVRL~\cite{richter2019open}, the first open-sourced reinforcement learning environment for surgical robotics, and showed that the learned policies can be deployed to the dVRK platform. However, it has not included an interface for physical input device, therefore could not support manual inputs through intuitive interactions. This further restricts the functionality on collecting human demonstration data to study machine learning algorithms. Tagliabue \emph{et al.} developed UnityFlexML~\cite{tagliabue2020soft} for effective RL policy learning on tasks involving soft-tissue manipulations. It includes a small number of sheet-like virtual assets to date, which may limit the diversity of surgical tasks or scenarios that can be investigated.
Very recently, Munawar \emph{et al.} released a comprehensive simulation platform AMBF~\cite{munawar2022open} which supports interactive human inputs, therefore it can collect data to train and test
control mechanisms for surgical robot. 
However, AMBF is not yet sufficiently supporting AI algorithmic libraries to facilitate people to explore reinforcement learning methods using large-scale human demonstrations.

In parallel with the current defect in simulators, 
a key unsolved technical pursuit of surgical robot learning is how to pursue a higher level of autonomy~\cite{yang2017medical,saeidi2022autonomous}. The solution to this problem needs to note that most existing surgical robots still use a human-centered manner (i.e. teleoperation)~\cite{d2021accelerating,zhang2019handheld}. In other words, AI developments should incorporate human factors into the loop in order to make use of expert's knowledge for promoting cognitive ability to learn complex surgical skills~\cite{zhang2022human}. 
However, achieving human-in-the-loop surgical embodied intelligence involves many challenges. First is how to establish an accurate and intuitive action mapping mechanism between human input device and surgical robot, which should be standardized and compatible for machine learning. Second is how to build realistic physical simulation accompanied with high-fidelity scene visualisation to provide immersive feedback from the virtual environment upon human interactions. Third is how to understand the effect of human demonstrations on policy learning for task automation in the context of surgical embodied AI.


To address above challenges, we propose to investigate human-in-the-loop embodied intelligence with an interactive simulator dedicated for surgical robot learning, with the concept illustrated in Fig.~\ref{TrainingFramework}. 
A human (usually an expert surgeon) can manipulate virtual objects via virtual robot arms through interaction with a physical input device, and in turn perceive visual feedback on such interactions in the simulator.
The human demonstrations can be saved to a database in the form of pairs of end-effector actions (variation of position, orientation and gripper angle) and environmental states, which are then used for control policy learning with RL. Besides, the robot can also explore the environment by itself through trial-and-errors (i.e. action-state pairs) in a typical embodied AI paradigm. 

In this work, to achieve the goal, we develop the platform based on our existing open-source simulator SurRoL~\cite{xu2021surrol}, i.e., a RL-centered and dVRK compatible surgical robot simulator. The proposed new version highlights the introduction of human interaction which is achieved by an interface of haptic input devices for two hands. Importantly, a set of new features are co-developed to establish the realistic human interaction: 1) standardized interface which supports human interaction through physical input device and policy learning, 2) realistic physical simulation with fine-grained modeling of collision properties and proportional derivative (PD) control, 3) high-fidelity rendering with vivid phong shading, spotlight modeling and shadow projection for human perception. On top of these, we further use collected human demonstration data to train policies for surgical robot task automation, and compare its performance with previous code-generated demonstration data. Promising results are achieved in terms of efficiency for control policy learning. Finally, five new surgical training tasks (i.e., \textit{PickAndPlace, PegBoard, NeedleRings, MatchBoard, MatchBoardPanel}) are added into the simulator which can support surgical skill training of practitioners. With these tasks, we aim to facilitate the future investigation of surgical robot learning on complex training tasks, and human-in-the-loop surgical embodied AI. Our code is available as an updated branch within the SurRoL repository at: \url{https://github.com/med-air/SurRoL}. 

\begin{figure*}[t]
    \centering
    \includegraphics[width=0.9\hsize]{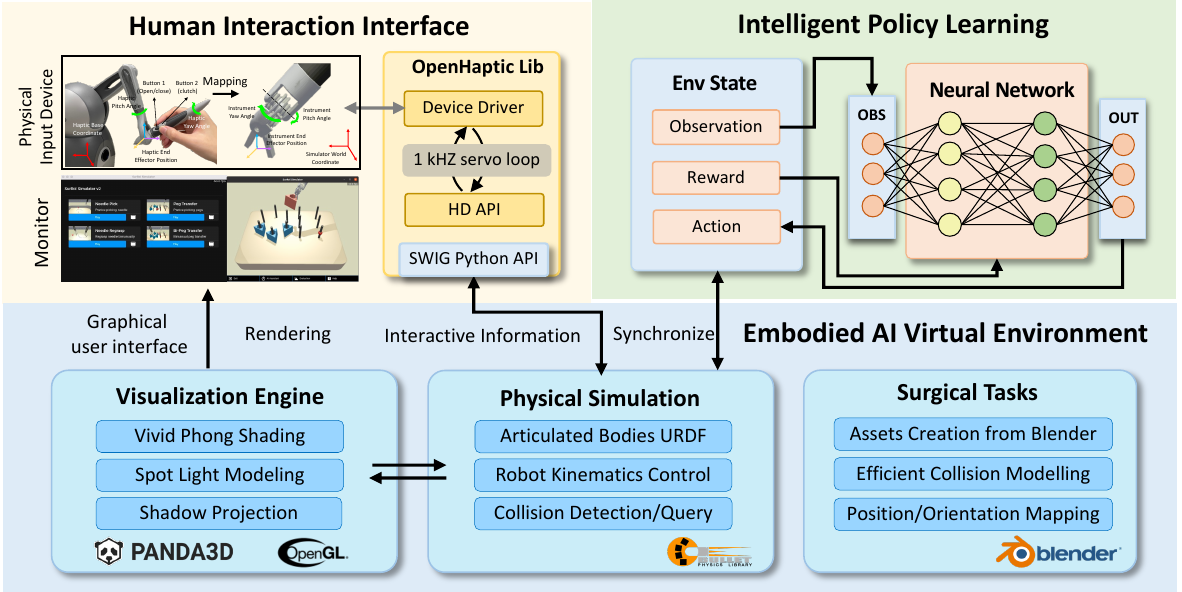}
    \vspace{-1mm}
    \caption{The proposed platform of human-in-the-loop surgical embodied intelligence with interactive simulation environment for surgical robot learning.}
    \label{Fig::overview}
    \vspace{-5mm}
\end{figure*}

\section{Related Work}
\subsection{Embodied AI for Robots with Interactive Simulators}
Embodied AI aims to learn complex skills through interaction with the environment, instead of directly learning from pre-collected datasets counting video, audio, image and text. Its rapid progress requires and promotes the development of simulators for embodied AI~\cite{duan2022survey}.
Several simulators with noticeable contributions from embodied AI community all incorporate interaction as an important feature. Xiang \textit{et al.} proposed SAPIEN~\cite{Xiang_2020_SAPIEN}, which enables robotic interaction tasks with a realistic and physics-rich simulated environment. Kiana \textit{et al.} proposed ManipulaTHOR~\cite{Ehsani2021ManipulaTHORAF} to facilitate research on visual object manipulation using a robotic arm. Recently, embodied AI with human interaction is gaining more and more attention. Li \textit{et al.} proposed iGibson~\cite{li2021igibson} to simulate multiple household tasks which allows VR-based human-environment interaction. Gan \textit{et al.} proposed ThreeDworld~\cite{gan2020threedworld} for interactive multi-modal physical simulation which supports human interaction through VR devices. Fu \textit{et al.} proposed RFUniverse~\cite{fu2022rfuniverse}, a physics-based action-centric environment for learning household tasks, which also provides VR interface for human input. Some studies~\cite{mandlekar2021matters,ramrakhya2022habitat} show that with human data involved, the intelligent algorithms tend to demonstrate better performance. 
Still, there are few dedicated attempts on simulators in surgical robot learning research topic, due to major challenges on developing open-source software infrastructures which support high-quality human and multiple surgical tasks for study.

\vspace{-0.5mm}
\subsection{Learning-based Surgical Task Automation}

Learning-based surgical task automation is an emerging field in surgical robotics that aims to automate surgical tasks using machine learning techniques. Various surgical tasks have been studied in this field, including suturing~\cite{schwaner2021autonomous}, pattern cutting~\cite{thananjeyan2017multilateral}, tissue manipulation~\cite{d2022learning}, etc. One of the commonly used approaches is imitation learning~\cite{hua2021learning}, where a policy is trained on a dataset of expert demonstrations through supervised learning technique. However, this method cannot handle distribution shift thus suffering from poor generalizability. On the other hand, RL~\cite{yip2019robot, scheikl2022sim}, which allows the robot to learn by interacting with the environment, shows outstanding capability of generalizing to learn different tasks. However, traditional RL will suffer from a large exploration burden, which is time-consuming and resource-intensive with no guarantee of success. Recently, there has been a growing interest in leveraging demonstration to improve the learning efficiency of RL~\cite{tan2019robot,pore2021learning,li20223d}, which achieved promising results. This type of methods are commonly studied with an interactive simulation environment for learning and testing due to the potential risks associated with real surgical procedures and the difficulty in obtaining comprehensive and accurate~\cite{richter2021robotic} surgical data. However, much developing workload is always accompanied to establish a simulator. 
In the light of this, we are dedicated to developing a open-sourced surgical robot simulator supporting high-quality human interaction and multiple tasks. With which we hope to accelerate progress in this rapidly evolving field.

%

\section{Materials and Methods}
\subsection{Simulation Platform Development as Infrastructure}
Our overall system consists of three main components: 1) human interaction interface, 2) intelligent policy learning, and 3) surgical embodied AI virtual environment. The overview framework is illustrated in Fig.~\ref{Fig::overview}. First, we create assets of surgical skill training tasks with the help of the 3D modeling tool Blender, and then generate relevant collision models and URDF description models at the same time for physical modeling in the simulation environment. Once surgical tasks are imported to the simulator, the human can conduct surgical action via a manual interaction device, and the interaction information is streamed to the virtual environment for physical simulation. In the meanwhile, the video frames are produced using the visualization engine of the simulator, which will be displayed on the monitor for human perception and next-step interaction. The policy can learn through interaction with the virtual environment by itself and also learn from human through recorded expert demonstration. 

\subsection{Kinematics Mapping and Control with Input Device}
To incorporate human control of surgical robots in the simulator, the first and essential step is to develop a manipulation system with physical input devices. 
In this paper, we opt for \textit{Touch}~\cite{arteaga2022geomagic} (3D Systems Inc.) as the typical input devices of the simulator owing to its advantages of high stability, customizability, wide adoption~\cite{munawar2022virtual,RoSS} and low cost. 
Specifically, two \textit{Touch} devices are used to simulate the two master arms of the robot to teleoperate the Patient Side Manipulators (PSMs) and the Endoscopic Camera Manipulator (ECM).


\subsubsection{Kinematics mapping from input device to virtual robot} 
To enable smooth control between the physical input device and surgical robots in the simulator, we map the current action of the end-effector instead of pose or joint angle from input device to the simulator, as it is more intuitive and compatible with the policy action (such as that from RL policy), which can also facilitate the recording of human demonstrations. In specific, for each simulation step $k$, we first retrieve the end-effector's position and joint angle of current step  $p(k) = \{x(k),y(k),z(k),rotate(k)\}$ and its previous step $p(k-1)$ from haptic (as shown in Fig.~\ref{Fig::overview}, physical input device). Then, we calculate the current action as $p(k)-p(k-1) = \{d_x,d_y,d_z,d_{rotate}\}$, where $d_x, d_y, d_z$ determine the position movement in the Cartesian space, $d_{rotate}$ determines the orientation change. For PSM, $d_{rotate}$ is yaw or pitch angle for top-down or vertical space setting, which can be adapted to meet specific surgical needs. For ECM, $d_{rotate}$ is the roll angle which allows the surgeon to adjust the camera's angle around its longitudinal axis. When Button 1 (shown in Fig.~\ref{Fig::overview}, physical input device) is pressed, the angle of instrument jaw $j(k)$ decreases a constant value for each step until it is closed ($j(k)<0$), while it is released, the $j(k)$ increases until it is fully open. When Button 2 is pressed, all actions are set to null to simulate the clutch mechanism of the control, which is usually used to adjust the master workspace during the operation. All movement actions will be multiplied by scaling vectors (corresponding to the tool movement scale), which will then be added to the current state of the surgical instrument to yield the target pose. 

\subsubsection{Instrumental end-effector control} After the action mapping, we realize the surgical robot control with inverse kinematic (IK) and a PD controller. Specifically, given the target pose of the surgical instrument end-effector, the target joint angles of the surgical robot are calculated using the IK based on Denavit-Hartenberg (DH) parameters (which are consistent with dVRK~\cite{d2021accelerating}). As the PSMs and ECM are respectively 6 Degree-of-freedom (DOF) and 4 DOF with a remote center of motion (RCM), the analytical solutions for both of them exist. Given the target and current pose of the instrument, we enable the realistic and smooth movement control of the robot using position and velocity PD control. The error is designed as:
\begin{equation}
\resizebox{.93\hsize}{!}{$e(t_k) = K_1\cdot(p_\text{target}(t_k)-p_\text{current}(t_k))+K_2\cdot(v_\text{target}(t_k)-v_\text{current}(t_k)),$}
\end{equation}
And the discrete control system~\cite{ang2005pid} is formulated as:
\begin{equation}
 \resizebox{.91\hsize}{!}{$u(t_k) = \left(K_p+\frac{K_d}{\Delta t}\right)e(t_k)-\left(K_p+\frac{2K_d}{\Delta t}\right)e(t_{k-1})+\frac{K_d}{\Delta t}e(t_{k-2}),$}
\end{equation}
where $u(t_k)$ represents the system output action, $K_1$ and $K_2$ stand for proportional and integral gains, and $\Delta t$ is the time interval. To maximize the efficiency of high-frequent controlling communication between haptic input and simulator, we leverage SWIG~\cite{beazley1996swig} to directly wrap the HD API of OpenHaptics (original Haptic SDK which is implemented using C/C++) into Python to bridge the haptic device with our python-based environment.


\subsection{Realistic Physical Interaction Simulation}

Introducing human interactions into the simulator will also introduce some unexpected actions (e.g., sudden movement and destructive behavior). In this regard, we need to further optimize the physical simulation and interaction based on SurRoL for a realistic manipulation user experience.

\begin{figure}[tp]
    \centering
    \includegraphics[width=1.0\hsize]{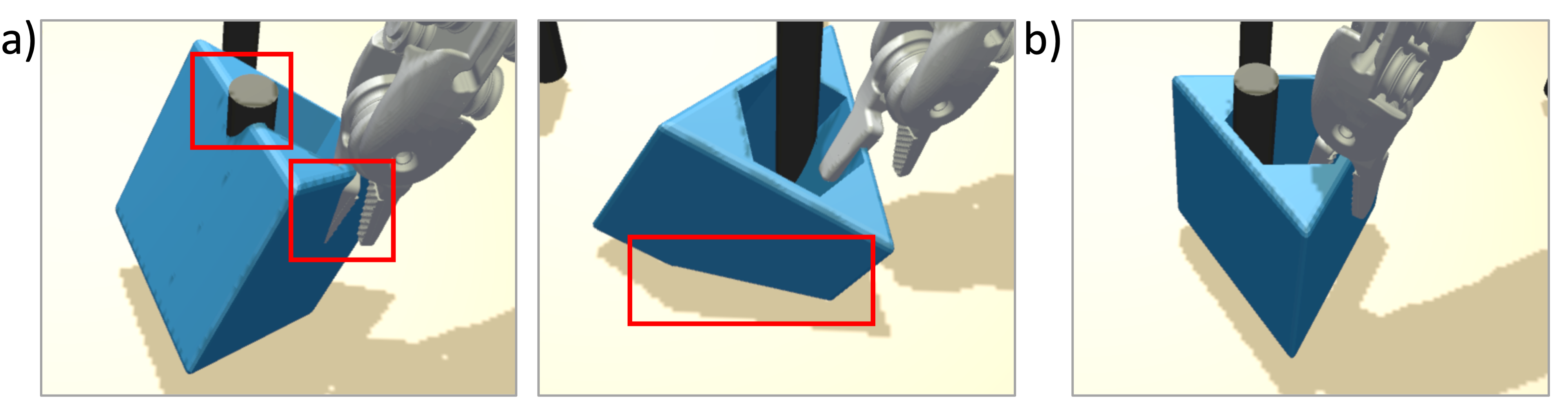}
    \vspace{-7mm}    
    \caption{The example results (a) before and (b) after optimizing physical modeling in \textit{PegTransfer} task (from SurRoL). It shows our proposed method can prevent inter-penetration problem under unexpected movement, yielding more realistic interaction when human manipulating in the simulator.}
    \label{Fig::physical}
    \vspace{-5mm}
\end{figure}

Firstly, to enable realistic simulation of different objects including the surgical instruments, all their articulated bodies are modeled following the real-world ratio using Blender. Meanwhile, the contact and inertia attributes (e.g. stiffness and mass) in objects' URDF files are first initialized with the real-world measurements and then adjusted manually through trail-and-error simulation. The adjustments are repeated iteratively until the simulation results match the real-world as closely as possible. Besides, given that most object assets in surgical tasks are more complicated than a single convex hull or primitive (e.g., boxes, cylinders, and spheres) \cite{munawar2022open}, convex decomposition by V-HACD \cite{vhacd} and collision primitive compound are applied to the meshes to get collision geometry in objects' URDF files for more precise collision detection.
However, when there are some destructive behaviors from humans, such as pushing or grasping objects with a very large motion, the inter-penetration problem could happen among surgical instruments and objects, which hurts realism. This problem arises when assets collide with each other with abnormally large speed, causing collision solving failed and making them intersected or overlapped. Moreover, the sudden movement of the tool from one place to another can not be achieved in real surgical robots. Therefore, designing the controller with constraints on force and velocity is of vital importance. Specifically, based on position and velocity PD control mode, we impose constraints on the output of joint motors when controlling PSMs and ECM. During the step simulation, the underlying designed control algorithms will calculate motor actions under the maximum motor force and velocity limitation to reach the target position. In addition, joint constraints are also utilized in the interaction between grippers and objects for more stable and realistic grasping. An example comparison of before and after optimization of the physical simulation is illustrated in Fig.~\ref{Fig::physical}, which shows our proposed solutions can prevent inter-penetration problem, yielding more realistic interaction.

\subsection{High-fidelity Surgical Scene Rendering}

\begin{figure*}[tp]
    \centering
    \includegraphics[width=0.95\hsize]{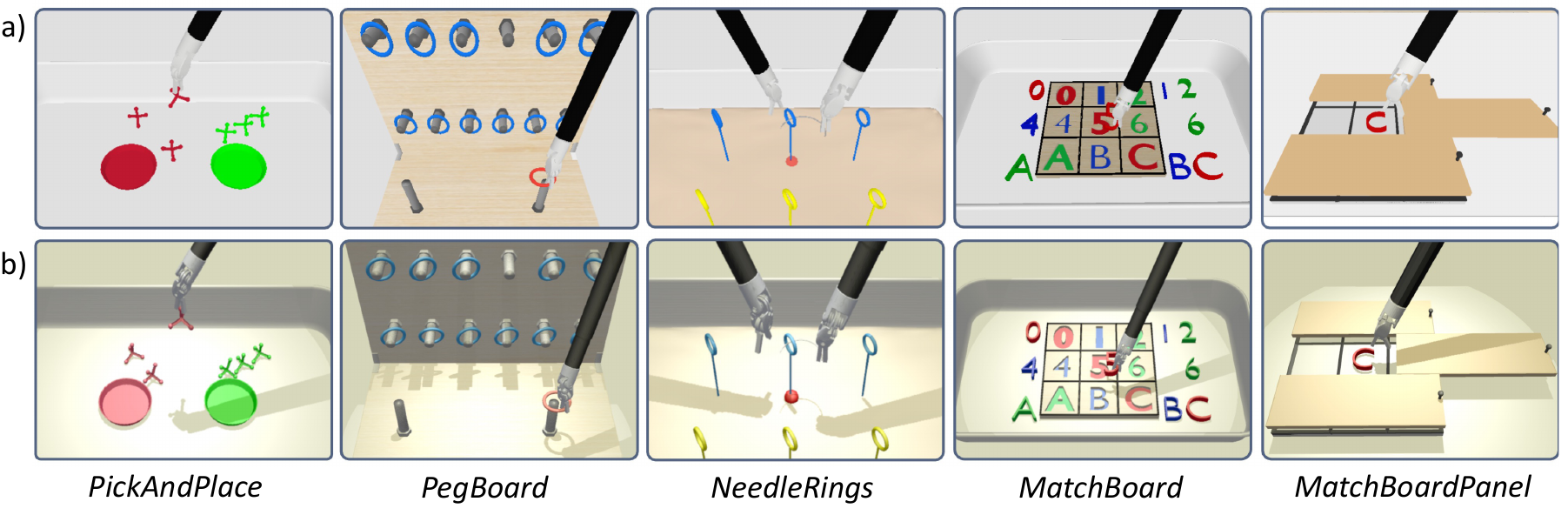}
    \vspace{-3mm}
    \caption{The visualization results (a) using original rendering engine from SurRoL, and results (b) using optimized engine for realistic rendering. Five tasks from left to right are 
    \textit{PickAndPlace}, \textit{PegBoard}, \textit{NeedleRings}, \textit{MatchBoard} and \textit{MatchBoardPanel} respectively, which are described in Sec.~\ref{Section::tasks}.}
    \label{Fig::render}
    \vspace{-5mm}
\end{figure*}

To develop an authentic human interaction in the simulator, providing the rendered scene with high visual realism is important for for human perception in virtual-reality based surgical training. However, the physical simulation backend in original SurRoL (using PyBullet) does not focus on rendering realism for human interaction thus only supports simple scene rendering for result visualization purpose. To address this limitation, we proposed a practical rendering plug-in for PyBullet engine which supports Panda3D~\cite{goslin2004panda3d} (an open source framework for 3D rendering with Python API). Our developed plug-in bridges the gap between PyBullet and Panda3D to enable realistic rendering of scenes.

Specifically, we rewrote the underlying rendering interface of PyBullet so that it can support conversion of the properties of assets, such as pose, texture, and material from PyBullet data format to Panda3D readable format, and then pass them to Panda3D for rendering. 
We adopt the phong shading~\cite{bishop1986fast} model from Panda3D to simulate the lighting reflection and diffusion of the whole virtual environment. Compared to the 
traditional rendering pipeline~\cite{gouraud1971continuous} using texture, normal maps and specular maps to control the realism of the scene, the adopted one is more similar to the physics of the real world. Moreover, the traditional way of simulating lighting sources is using directional light or ambient models. They are inconsistent with the endoscope light which adopts fiber optic cable to guide external light sources to the end of the endoscopy and emit an intensive cone beam directionally toward the area of interest (surgical scene). To simulate this kind of lighting condition, the spotlight model, which can replicate this effect by focusing the light in a particular direction with field-of-view, is leveraged to achieve a realistic simulation of surgical robot endoscope lighting. In addition, shadow mapping is enabled to simulate the light occlusion and visualize the projected shadow. Last but not least, the previous rendering engine will produce images with an obvious aliasing effect which can affect the human perceptual experience to a large degree. As a result, we enable the anti-aliasing in the rendering pipeline for a more natural visualization. The comparison of the visualization results is shown in Fig.~\ref{Fig::render}, which demonstrates the proposed method can generate more realistic visualizations in terms of lighting, shadow, reflection and fidelity. Finally, in the spirit of user-friendliness, we further leverage Panda3D to develop an graphical user interface (GUI) (as illustrated in Fig.~\ref{Fig::overview}, Monitor). It allows trainee to conveniently select different surgical training task in a panel through clicking the ``play" button, and exit the task through ``exit" button, which can be flexibly further extended and customized as needed.

Notably, the proposed pipeline can not only improve the rendering realism for human interaction, but also can largely facilitate the further research on image-based perception and sim-to-real tasks, such as visual reinforcement learning~\cite{scheikl2022sim}, surgical scene segmentation and action recognition, bridging the gap between the virtual environment and the real world. 

\subsection{Surgical Tasks for Both Training and Automation}
\label{Section::tasks}
In the previous version of SurRoL simulator, we designed several surgical robot tasks (e.g., \textit{NeedlePick}, \textit{NeedleRegrasp} and \textit{EcmReach}) to specifically evaluate the robot learning algorithms. Still, these tasks are not sufficient to fulfill the requirements of current and future comprehensive research on surgical training~\cite{9807505,smith2014fundamentals}, where human interaction is a very important factor. To this end, we add five new tasks following the common curriculum tasks in robotic surgery simulation-based training~\cite{sanford2022association,perrenot2012virtual,cowan2021virtual}, which are representative of the evaluation of both surgical training and task automation. Specifically, we add new tasks of \textit{PickAndPlace}, \textit{PegBoard}, \textit{NeedleRings}, \textit{MatchBoard} and \textit{MatchBoardPanel} (see Fig.~\ref{Fig::render}). 
In task \textit{PickAndPlace}, the trainee needs to pick up the colored jacks and place them in the tray in the same color. The task will be considered as success only when all the colored jacks are placed on the corresponding trays.
In task \textit{PegBoard}, the trainee needs to pick up the ring from the vertical peg board and then place it on the peg from the horizontal board, which requires high proficiency in controlling the pose and orientation of the rings thus effective for trainees to practice their manipulation skills. In task \textit{NeedleRings}, the trainee needs to hand off the needle using two robot arms to pass through the ring, which calls for trainees to master proficient two-handed needle manipulation skills in terms of precise position and orientation control. 
In task \textit{MatchBoard}, the trainee needs to pick up a digit or alphabet block and place it on groove in the corresponding position, which further practices hand-eye coordination. 
In task \textit{MatchBoardPanel}, the trainee needs to first grasp the drawer handle to open the drawer and then pick up a digit or alphabet block to place it in the targeted grid, which is an advanced skills training tasks with sequential sub-tasks.
These tasks are specifically designed and modeled which can be easily extended and applied for further research on surgical training and human-in-the-loop robot learning.


\subsection{Surgical Skills Learning from Human Demonstrations}
While traditional methods of learning from demonstration often involve exploiting data and knowledge from machine-generated demonstrations~\cite{levine2013guided,vecerik2017leveraging}, such approaches may not be effective for surgical tasks, which require delicate and skillful surgical operations by surgeons that cannot be interpreted through straightforward actions. Therefore, in the case of robotic surgery, the primary focus of learning from demonstration is on expert human demonstrations~\cite{su2021toward,li20223d}. In this regard, we opt for human demonstration-guided RL as example to validate our proposed interactive simulator.

\subsubsection{Formulation}
Specifically, we consider an RL agent interacts with our simulator formulated as a Markov Decision Process. The agent takes an action $a_k\in{\mathcal{A}}$ at state $s_k\in\mathcal{S}$ according to its policy $\pi:\mathcal{S}\rightarrow\mathcal{A}$ at each time step $k$. It then receives a reward signal $r_k$ and transits to the successor state $s_{k+1}$, repeating policy execution until the episode ends, where each experience $(s_k,a_k,r_k,s_{k+1})$ is stored into a replay buffer $\mathcal{D}_A$. Meanwhile, the agent maintains another buffer $\mathcal{D}_{E}$ that includes expert demonstrations given in advance. The goal of the agent is to find an optimal policy $\pi^\star$ that maximizes the expectation of discounted cumulative reward (a.k.a. return) with a discount factor $\gamma\in(0,1]$. Many RL methods achieve this by estimating Q-value function $Q:\mathcal{S}\times\mathcal{A}\rightarrow \mathbb{R}$ that gives the expected return of action $a_k$ at state $s_k$. We herein adopt deep deterministic policy gradient DDPG~\cite{lillicrap2015continuous}, a widely-used deep RL method in surgical robot learning~\cite{richter2019open,d2022learning}, to learn a Q-value function $Q_\theta$ with parameters $\theta$ by minimizing the squared Bellman error~\cite{lillicrap2015continuous}:
\begin{equation}
 \resizebox{.91\hsize}{!}{$L_Q(\theta) = \mathbb{E}_{\mathcal{D}_A}\left[ \left(r_k+\gamma Q_\theta(s_{k+1},a_{k+1}) - Q_\theta(s_k,a_k)\right)^2\right].$}
\end{equation}

To exploit the knowledge from demonstrations,
we follow the approach in~\cite{goecks2020integrating} that utilizes state-action pairs from demonstrations to encourage the behavioral similarity between agent and expert. It realizes this by first pre-training the policy $\pi_\phi$ parameterized by $\phi$ with behavior cloning loss~\cite{bain1995framework}, and then minimizing the following objective at the online stage:
\begin{equation}
\resizebox{.95\hsize}{!}{$\mathcal{L}_\pi(\phi) = \mathbb{E}_{{s}\in\mathcal{D}_A}\left[-Q_\theta(s_k,\pi_\phi(s_k))\right] + \mathbb{E}_{{s,a}\in\mathcal{D}_E}\left[ \|\pi_\phi(s_k)-a_k\|_2^2 \right].$}
\end{equation}

We further adopt hindsight experience replay HER \cite{andrychowicz2017hindsight} as a sampling strategy for both buffers, which addresses the sparse reward issue for goal-conditioned environments.

\subsubsection{Experiment setup}
We choose \textit{NeedlePick} as a representative task for validation, which is an essential surgical training task, where the user needs to manipulate the robot arm to reach and pick up the needle on the tray, and then move it to a targeted location. It contains multiple interactions with the virtual environment thus relevant surgical skill is needed to successfully complete the task. 
The reward function is defined following the previous SurRoL as: $r(s,a) = -\mathbb{I}_{({\left\|o_{g}-o_{c}\right\|_2}\geq\epsilon)}$,
where $o_{c}$ and $o_{g}$ are respectively current and goal location of the needle center, while $\epsilon$ is the tolerant distance between them.
During the experiment, we collect the successful human expert and non-expert demonstrations using the proposed simulator. Meanwhile, we also developed a linear path planning script which can generate a sequence of waypoints between the robot's current position and the target position to collect script demonstrations. Then we compare the results of using human expert, non-expert demonstrations and the result of using script demonstrations. To evaluate the learned policies, we opt for success rate, steps to complete (time cost), and economy of motion (trajectory distance) as the evaluation metrics, which have been commonly adopted to evaluate the surgeons' skills in the robotic surgery simulation training~\cite{perrenot2012virtual,whittaker2016validation,cowan2021virtual}, owing to their effectiveness of representing the level of surgical skills.

\subsubsection{Implementation details}
Following the setting in SurRoL, the workspace is set to $10.0cm^{2}$, and the tolerance distance between the goal and current state is $0.5cm$. Each training epoch contains 40 episodes and each episode is set to 150 timesteps. Simulation time step is set to $0.2$ seconds. The environment will be reset if the user fails within the defined timesteps. The initial and goal conditions are randomly sampled every time the environment resets. The policy is modeled as four-layer MLPs with ReLU activations, where each layer was of 256 hidden dimensions. We build our implementation using DEX~\cite{huang2023guided} framework, which is on top of OpenAI baselines~\cite{baselines} and use their hyper-parameter settings for fair comparison. For demonstration collection, we record 100 human expert demonstrations from four medical students (each 25), who had conducted robotic surgical training and fluent at using our simulator. In addition, we record 100 non-expert human demonstrations from four engineering students (each 25), who had hands-on experience using our simulator. Meanwhile, we generated additional 100 script demonstrations through our designed path planning script for comparison.
For fairness, the average completion time steps of demonstrations are numerically close with around 113 steps for the human expert, 117 for non-expert and 109 steps for the script. Similarly, the average trajectory distances are $12.1cm$ for human expert, $12.5cm$ for non-expert and $11.7cm$ for script demonstrations.


\section{Results}

Current scheme of evaluating surgical simulation training through proficiency-based progression (PBP)~\cite{satava2019future,gallagher2011fundamentals} emphasizes the importance of learning efficiency to assess the level of surgical skills, which parallels the concept in previous works on RL~\cite{laskin2020curl,laskin2020reinforcement} that advocate for evaluating the sample-efficiency of proposed algorithms at the early stage of training. In the light of this, to analyze the learning process and policy performance, we analyse the training curve of the policy and the results at 50 epochs (early training stage) and 100 epochs (when all the results show no increasing trend).

\begin{figure}[b]
    \centering
    \vspace{-4mm}
    \includegraphics[width=0.85\hsize]{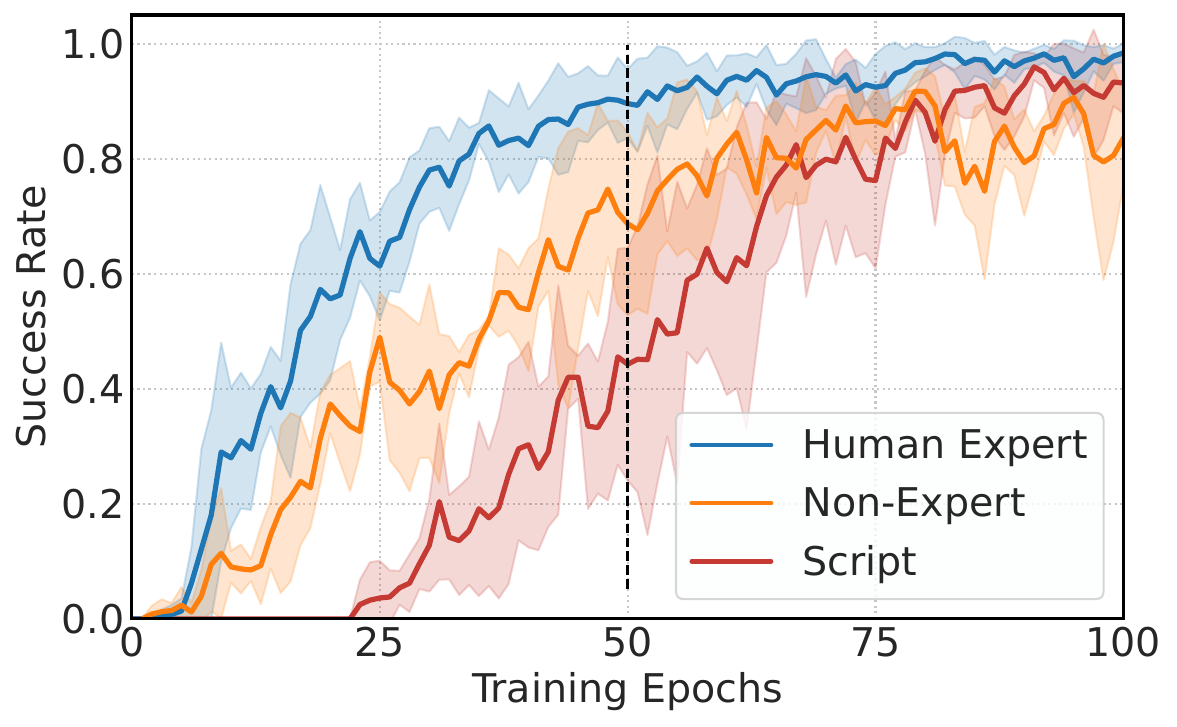}
    \vspace{-4mm}
    \caption{The success rate learning curve for \textit{NeedlePick} with a sliding window smooth. Results show that policy learned from human expert demonstrations consistently outperforms the policy learned from non-expert demonstrations or script demonstrations. The shaded region represents the standard deviation over four random seeds.}
    \label{Fig::resultCurve}
    \vspace{-1mm}
\end{figure}

\begin{table*}[t]
\caption{
\textbf{The evaluation results of \textit{NeedlePick} in simulator.}}
\centering
\vspace{-2mm}
 \resizebox{1.7\columnwidth}{!}{
\begin{tabular}{c|ccc|ccc}
\hline
\multirow{2}{*}{Types of Demo } & \multicolumn{3}{c|}{50 epochs} & \multicolumn{3}{c}{100 epochs} \\ \cline{2-7} 
                            & ~~Script    & Non-expert~~             & Human Expert~~~                     & ~~Script   & Non-expert~~   & Human Expert~~~     \\ \hline
Success Rate / \% ($\uparrow$)               & ~~~43.2\ci{21.9}  &   66.3\ci{12.7}      & \textbf{89.7\ci{6.4}}   &  ~~~93.2\ci{6.9}     &  86.7\ci{2.4}     &  \textbf{99.3\ci{0.1}}     \\
Steps to Complete ($\downarrow$)        & ~~~101.8\ci{27.1} &   59.5\ci{17.1}      &\textbf{39.2\ci{14.5}}  &  ~~~24.0\ci{10.4}     &  21.4\ci{5.9}    &  \textbf{15.0\ci{5.5}}   \\
Economy of Motion / cm ($\downarrow$)   & ~~15.0\ci{2.5}   &     14.2\ci{3.1} &\textbf{12.2\ci{1.3}}  &  ~~~13.5\ci{1.6}     &  12.6\ci{2.9}    &   \textbf{11.6\ci{1.1}}    \\ \hline
\end{tabular}
}
\label{table::results}
\vspace{-2mm}
\end{table*}

To demonstrate the learning efficiency of the policies, we train them using four seeds and plot the averaged success rates with results shown in Fig.~\ref{Fig::resultCurve}. We observe that learning from human expert demonstration consistently outperform learning from non-expert and script demonstrations in terms of the success rate over the training process. Specifically, the learning curves of RL policies from human expert demonstrations have smaller deviations than non-expert and script, indicating the improvement in stability of policy learning. After initial training for 50 epochs, the averaged success rate of \textit{NeedlePick} achieves $89.7\%\pm6.4\%$ when learning from the human expert, outperforming learning from the non-expert ($66.3\%\pm12.7\%$ success rate) by $23.4\%$, and outperforming script ($43.2\%\pm21.9\%$ success rate) by $46.5\%$ (as shown in Table~\ref{table::results}). The policy learned from human expert demonstration can also master a faster completion time steps ($39.2\pm14.5$) compared to non-expert ($59.5\pm17.1$) and script ($101.8\pm27.1$). Moreover, we calculate the average economy of motion and find that the motion trajectories are shorter with $12.2cm\pm1.3cm$ when using human expert demonstration, $14.2cm\pm3.1cm$ for non-expert and $15.0cm\pm2.5cm$ for script. After training for 100 epochs, the results of learning from human demonstration are more stable than non-expert and script with smaller variance. Meanwhile, the policy learned from human expert demonstration consistently outperformed the policy learned from non-expert or script for all metrics. Results demonstrate the improvement over the efficiency of policy learning gained from expert human demonstrations.

\begin{figure}[tp]
    \centering
    \includegraphics[width=1.0\hsize]{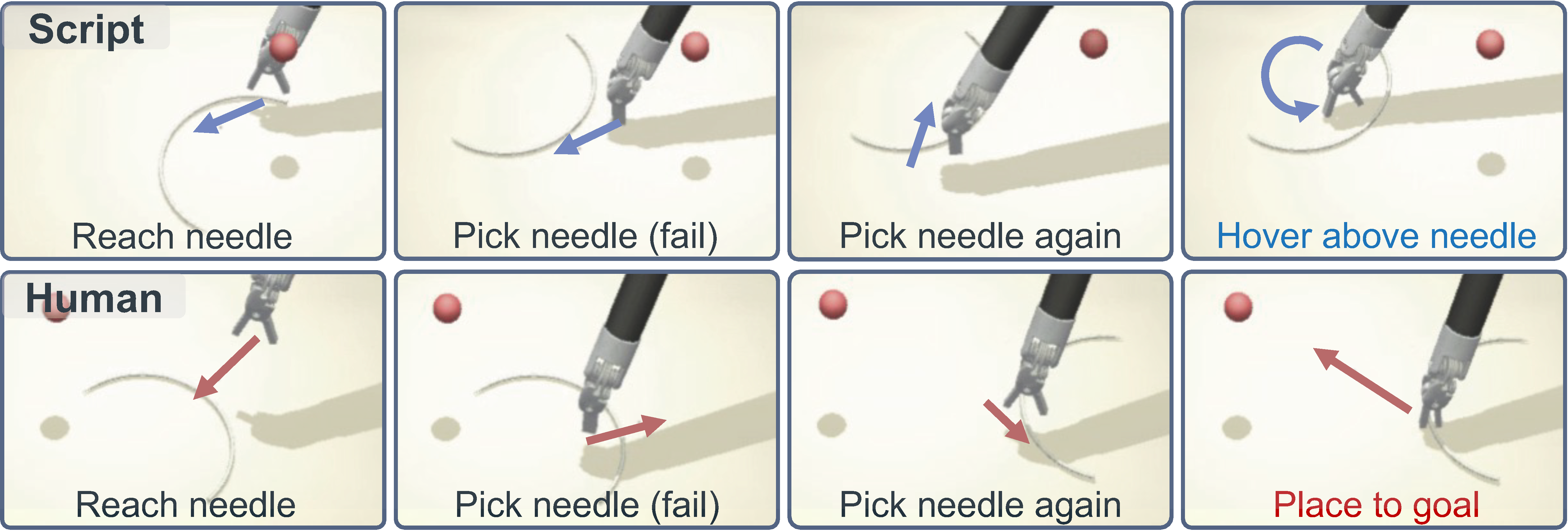}
    \vspace{-7mm}
    \caption{An example case shows that at early training stage, the policy learned from human demos can master skills of handling failure attempts and then complete the task, while learning from script demos cannot.}
    \label{Fig::resultTraj}
    \vspace{-5mm}
\end{figure}

Apart from the above quantitative results, we further analyze the qualitative results at 50 epochs. We visualize all the cases in the format of videos, and compare the actions from policies that were learned from the human (including expert and non-expert) and the script. A distinct observation has been found as illustrated in Fig.~\ref{Fig::resultTraj}. The first row shows a failure case of policy learned from the script, where the instrument first reaches the needle and tries to pick it up, but fails to grasp it. After failing on the first attempt, the policy could not grasp the needle again in the following steps. Instead, when learning from human demonstrations, the policy can grasp the needle again and complete the task after failing on the first failure grasping trail. Results show that using human demonstrations can help policy more quickly grasp the skills to handle failure cases at early learning stage, which demonstrate the effectiveness of incorporating the human factor for policy learning in the form of human demonstration.

\vspace{-1mm}
\section{Conclusion and Discussion}
\vspace{-1mm}
In this paper, we propose an interactive platform based on our existing SurRoL simulator for human-in-the-loop embodied intelligence. New features are added for enhanced human interaction, including haptic interaction interface, physical simulation and scene rendering with better realism. Five new representative surgical training tasks are also co-developed for future work on human-in-the-loop surgical task learning. We initially conduct basic experiments to validate the idea of including human in the embodied intelligence in the form of human demonstration. The promising results prove the effectiveness of our proposed methods and pave the way for relevant research topics. 
Potential future directions include taking the surgical safety~\cite{thananjeyan2020safety,pore2021safe} into consideration for policy learning from human demonstration, studying the effectiveness of learning from human feedback, and human-robot collaboration~\cite{zhang2022human}. We envisage extensive future works to explore how human factors can play a role and transform surgical robot learning through our provided open-source embodied AI platform for surgical robots.

{\footnotesize
\bibliographystyle{IEEEtran}
\bibliography{ref}
}

\end{document}